\documentclass[11pt]{article}
\usepackage[final]{acl}

\usepackage{adjustbox}
\usepackage{multirow}
\usepackage{multicol}
\usepackage{wrapfig}
\usepackage{scalerel}
\usepackage{float}
\usepackage[caption = false]{subfig}

\usepackage[T1]{fontenc}    % use 8-bit T1 fonts

\usepackage[utf8]{inputenc}

\usepackage[final]{microtype}

\usepackage{newtxtext,newtxmath} % <- this switches main font to TeXGyreTermesX*

\usepackage{svg}

\usepackage[scaled=1.02]{zi4}   % a hair larger (try 1.00–1.05)
\usepackage[htt]{hyphenat}      % allow hyphenation in \texttt
\usepackage{upquote}            % straight quotes in verbatim/\texttt
\microtypecontext{protrusion=basictext}

\usepackage[normalem]{ulem} % underlines
\usepackage{twemojis}
\usepackage{soul}
\usepackage{datetime}       % dates
\usepackage{booktabs}       % professional-quality tables
\usepackage{nicefrac}       % compact symbols for 1/2, etc.
\usepackage{xcolor}         % colors
\usepackage{url}            % simple URL typesetting
\usepackage{fnpct}           % nicer footnotes
\usepackage{enumitem}        % nicer itemize
\usepackage{tcolorbox}       % colored background
\usepackage{csquotes}        % quotation marks
\usepackage{xspace}          % adds whitespace after macros
\usepackage{graphicx}

\makeatletter
\xspaceaddexceptions{\csq@qclose@i}
\makeatletter

\usepackage[scaled=0.85]{helvet}

\usepackage{bbm}
\usepackage{amsfonts}

\usepackage{amssymb}
\usepackage{mathtools}

\usepackage{amsmath}
\usepackage{bm}
\usepackage{amsthm}
\usepackage{thmtools} 
\usepackage{thm-restate}
\usepackage{cancel}

\usepackage{nicefrac}

\usepackage[createShortEnv,commandRef=Cref]{proof-at-the-end}

\newtheoremstyle{resultstyle} % name of the style to be used
  {.8em} % space above
  {.8em} % space below
  {} % font in the body of the theorem
  {} % indent amount
  {\bfseries} % theorem head font
  {.} %  punctuation after theorem head
  {.5em} % space after theorem head
  {\thmname{#1 }\thmnumber{#2. }\textbf{\thmnote{#3}}}

\theoremstyle{resultstyle}

\usepackage{algorithm}
\usepackage{algpseudocode}  % requires algorithmicx
\makeatletter
\algnewcommand{\LineComment}[1]{\Statex \hskip \ALG@thistlm \textcolor{blue}{// #1}}
\algnewcommand{\FirstLineComment}[1]{\Statex \hskip\ALG@tlm \textcolor{blue}{\(\triangleright\) #1}}
\algnewcommand{\InlineComment}[1]{\hfill\textcolor{blue}{\(\triangleright\) #1}}
\makeatother

\usepackage{cleveref}
\crefname{section}{\S}{\S\S}
\Crefname{section}{\S}{\S\S}
\crefformat{section}{\S#2#1#3}
\crefname{figure}{Fig.}{Fig.}
\crefname{alg}{Alg.}{Alg.}
\crefname{line}{line}{lines}
\crefname{appendix}{App.}{App.}
\crefname{equation}{eq.}{eqs.}
\crefname{table}{Table}{Tables}
\crefname{proposition}{Proposition}{Propositions}
\crefname{assumption}{Assump.}{Assumps.}
\crefname{lemma}{Lemma}{Lemmas}
\crefname{definition}{Defn.}{Defns.}
\crefname{hypothesis}{Hypothesis}{Hypotheses}
\crefname{estimator}{Estimator}{Estimators}
\crefname{theorem}{Theorem}{Theorems}
\crefname{thm}{Theorem}{Theorems}
\crefname{result}{Result}{Results}

\usepackage{tabularray}
\usepackage{cellspace}
\newcommand\cincludegraphics[2][]{\raisebox{-0.4\height}{\includegraphics[#1]{#2}}}
\usepackage{setspace}

\usepackage{siunitx}
\DeclareSIUnit[quantity-product = {}, reset-math-version = false]\thousand{k}
\DeclareSIUnit[quantity-product = {}, reset-math-version = false]\million{M}
\DeclareSIUnit[quantity-product = {}, reset-math-version = false]\billion{B}
\DeclareSIUnit[quantity-product = {}, reset-math-version = false]\trillion{T}

\DeclareSIUnit[quantity-product = {}, reset-math-version = false]\x{x}
\DeclareSIUnit[quantity-product = {}, reset-math-version = false]\percent{\%}

\DeclareSIUnit[quantity-product = {}, reset-math-version = false]\hour{h}
\DeclareSIUnit[quantity-product = {}, reset-math-version = false]\min{m}
\DeclareSIUnit[quantity-product = {}, reset-math-version = false]\sec{s}

\xspaceaddexceptions{\textsubscript}

\makeatletter
\DeclareRobustCommand*{\escapeus}[1]{%
  \begingroup\@activeus\scantokens{#1 }\endgroup}
\begingroup\lccode`\~=`\_\relax
   \lowercase{\endgroup\def\@activeus{\catcode`\_=\active \let~\_}}
\makeatother

\usepackage{pifont}
\newcommand{\cmark}{\textcolor{green!60!black}{\ding{51}}} % green checkmark
\newcommand{\xmark}{\textcolor{red}{\ding{55}}}     
\newcommand{\warnmark}{\textcolor{orange}{\ding{115}}} % Triangle symbol

\title{
    \centering
    \raisebox{-0.45em}{\includegraphics[height=1.5em]{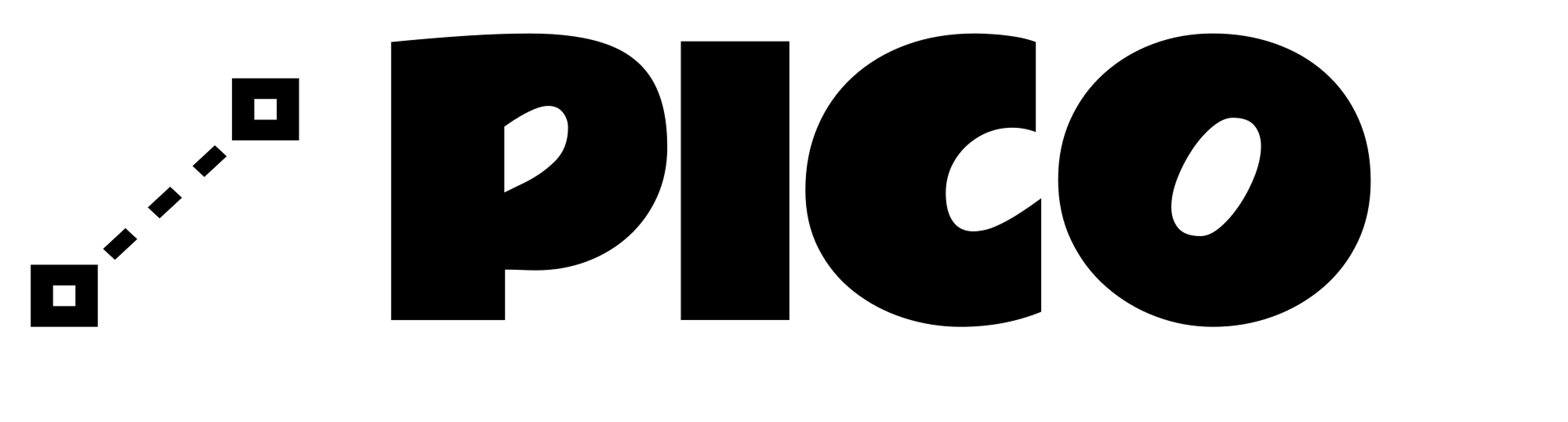}}A Modular Framework for Hypothesis-Driven Small Language Model Research
}

\author{
    \textbf{
        Richard Diehl Martinez\thanks{Corresponding author: \textbf{richard@picolm.io}} ~~~
        David Demitri Africa\thanks{Equal contribution} ~~~
        Yuval Weiss\footnotemark[2]} \\
    \textbf{
        Suchir Salhan ~~~
        Ryan Daniels ~~~
        Paula Buttery }\\
    University of Cambridge
}

\begin{document}
\maketitle
% =======================
% Abstract
% =======================
\vspace*{-1cm}
\begin{abstract}

%Building language models (LMs), especially small and medium-sized ones, remains more an art than a science. While large LMs can often improve simply by scaling up, it is still unclear why specific design choices work as they do. For small LMs, this uncertainty is even more limiting, since tight parameter budgets make each design decision critical to performance. Yet researchers still lack systematic, scientifically grounded ways to develop and test new ideas in this regime.

Building language models (LMs), especially small and medium ones, remains more art than science. While large LMs often improve by sheer scale, it is still unclear why many design choices work. For small LMs, this uncertainty is more limiting: tight parameter budgets make each decision critical, yet researchers still lack systematic, scientific ways to test and refine new ideas. We introduce \textbf{Pico}, a lightweight, modular framework that enables systematic, hypothesis-driven research for small and medium-scale language model development. \textbf{Pico} consists of two libraries that together provide a practical sandbox where researchers can make targeted changes to a model's architecture or training procedures and directly observe their effects on the model's behavior. To support reproducible experimentation, we also release a suite of baseline models, \textbf{\texttt{pico-decoder}}, trained under standardized conditions and open-sourced for the community. Case studies highlight how \textbf{Pico} can support iterative small LM design and analysis.

\begin{tblr}{
  colspec = {Q[c,m] X[l,m] Q[c,m] X[l,m]},
  rowsep = 3pt,
  stretch = 0
}
\cincludegraphics[width=1.9em]{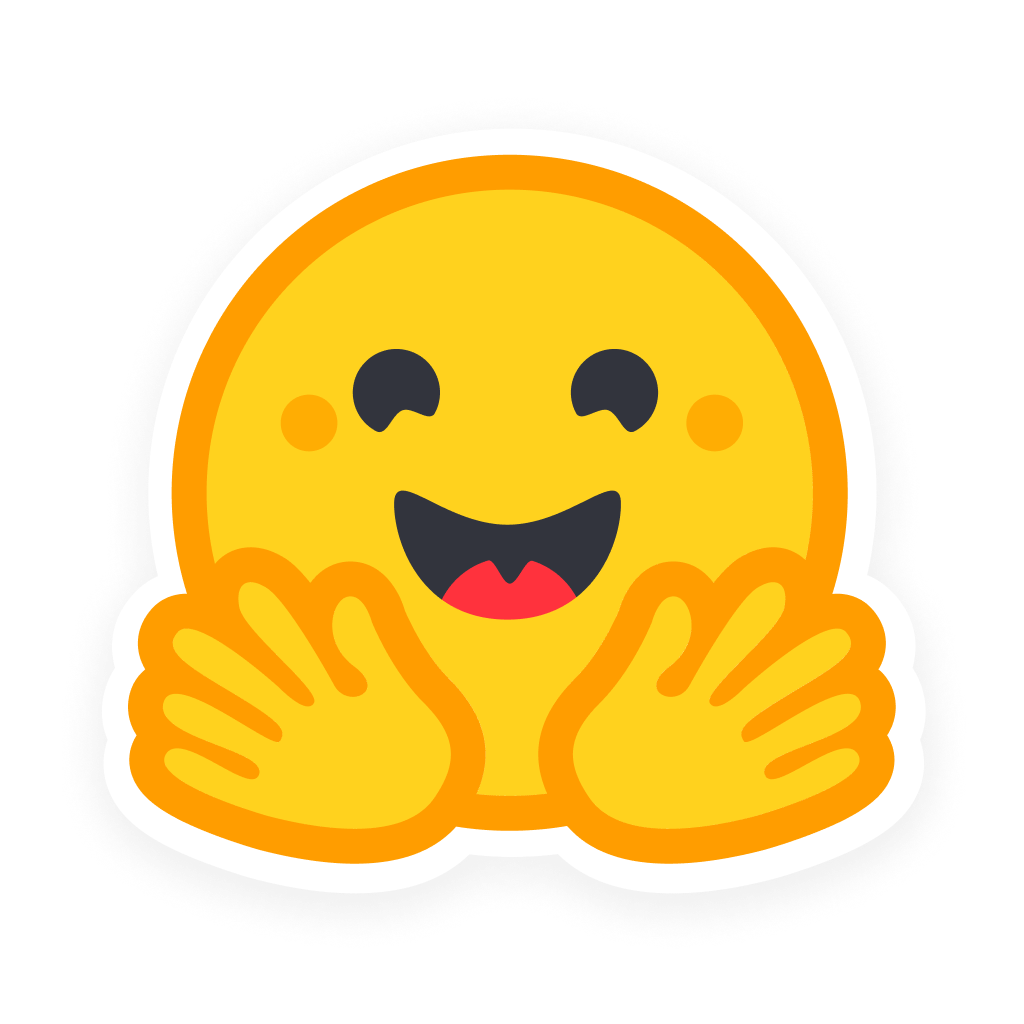} & 
{\footnotesize{\href{https://huggingface.co/pico-lm}{pico-lm} \newline \tiny{(Apache 2.0)}}} &

\vspace{-0.01cm}
\cincludegraphics[width=1.5em]{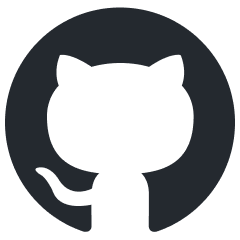} & 
{\footnotesize{ \href{https://github.com/pico-lm}{pico-lm} \tiny{(Apache 2.0)}}} \\

\cincludegraphics[width=1.4em]{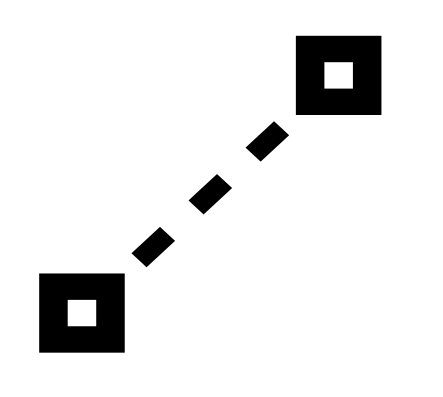} & 
{\footnotesize{\href{https://www.picolm.io/demo-paper}{picolm.io}}} &

\vspace{-0.01cm}
\cincludegraphics[width=1.5em]{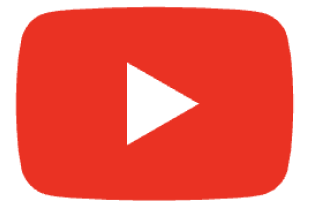} & 
{\footnotesize{\href{https://youtu.be/llRUKwqMah4?si=F4Ol8P5Tj2ZQB7Fm}{Demo Video}}}
\end{tblr}

\end{abstract}

% =======================
% The Body
% =======================

\section{Introduction}

{
\renewcommand{\arraystretch}{1.25}
\setlength{\tabcolsep}{4pt}

\begin{table*}[htbp]
    \centering
    \footnotesize
    \begin{tabular}{@{}p{2.4cm} p{1.7cm} p{2.4cm} p{2.3cm} p{2.2cm} p{2.2cm}@{}}
    \toprule
    \textbf{Tool} &
    \textbf{Custom \newline Training} &
    \textbf{Checkpoint \newline Support} &
    \textbf{Feature \newline Extraction} &
    \textbf{Analysis \newline Tools} &
    \textbf{Low-Budget \newline Friendly} \\
    \midrule
    \textbf{Pico} & 
    \cmark \newline Modular \newline PyTorch &
    \cmark \newline Optimizer, \newline weights \& data &
    \cmark \newline Activations \& \newline gradients &
    \cmark \newline \texttt{pico-analyze} \newline metrics &
    \cmark \newline Academic-GPU \newline scale \\

    \midrule

    TransformerLens & 
    \xmark & \xmark & \cmark & \cmark & \cmark \\

    ACDC & 
    \xmark & \xmark & \cmark & \cmark & \cmark \\

    SAELens & 
    \xmark & \xmark & \warnmark & \cmark & \cmark \\

    \midrule

    SmolLM2 & 
    \cmark & \warnmark & \xmark & \xmark & \cmark \\

    Pythia Suite & 
    \warnmark & \warnmark & \xmark & \warnmark & \cmark \\

    OLMo & 
    \cmark & \warnmark & \xmark & \xmark & \warnmark \\

    \bottomrule
    \end{tabular}

    \caption{Comparison of \textbf{Pico} and related frameworks for interpretability and learning dynamics. 
    \label{tab:pico_comparison} \newline
    \textbf{Legend:} \cmark = Fully supported; \warnmark = Partial; \xmark = Not supported.}
\end{table*}
}

\begin{figure}
    \centering
    \includegraphics[width=\linewidth]{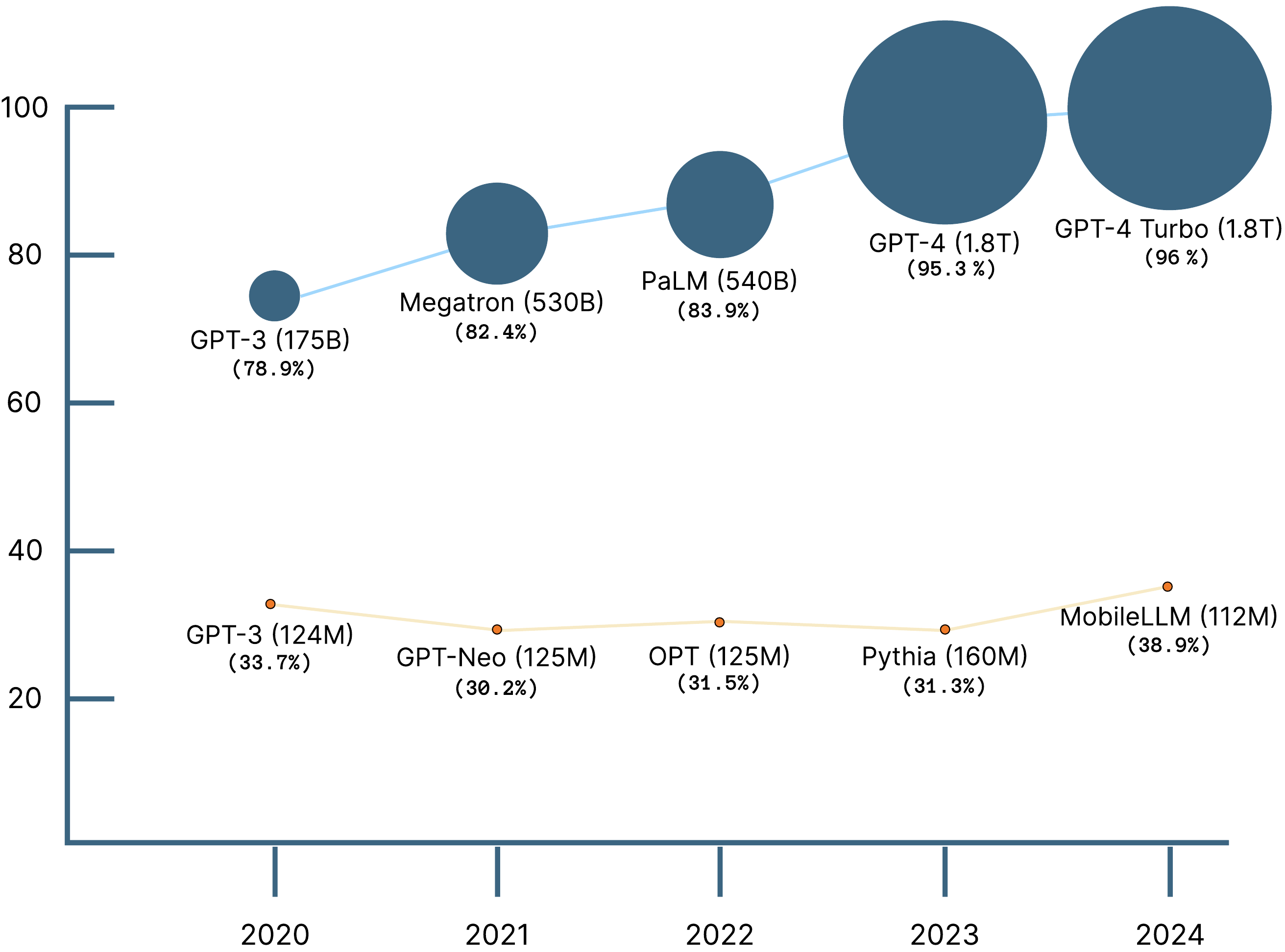}
    \caption{Best performance of fixed-size SLMs and LLMs on MMLU per year, 
    size of the circles represents model size. Note that the size of GPT-4 models is speculative\footnotemark{} 
    and has not been confirmed by OpenAI.}
    \label{fig:lm_performance_comparison}
\end{figure}

Recent advances in large language models (LLMs) have enabled strong performance across diverse tasks \citep{hendrycks2021mmlu, cobbe2021gsm8k, srivastava2023bigbench}, but progress on Small Language Models (SLMs) has been slower (see \cref{fig:lm_performance_comparison}). SLMs, loosely defined as models with fewer than 10 billion parameters, are large enough for emergent behaviors yet small enough to train on modest budgets \citep{hu2024minicpm, van2024survey, wang2024comprehensive, subramanian2025small}. Despite growing interest, designing efficient, high-performing SLMs still relies on opaque trial-and-error, with limited understanding of how design choices shape learning dynamics.

%Recent advances in large language models (LLMs) have produced systems capable of remarkable performance across a wide range of natural language tasks \citep{hendrycks2021mmlu, cobbe2021gsm8k, srivastava2023bigbench}. 

%Option 1 - SLMs framing 
%However, progress in designing Small Language Models (SLMs) has been much slower (see \cref{fig:lm_performance_comparison}). SLMs, though only loosely defined, usually have fewer than 10 billion parameters. This makes them large enough for specialized emergent behaviors to arise, yet small enough to train and study on modest compute budgets. \citep{hu2024minicpm, van2024survey, wang2024comprehensive, lu2024small, subramanian2025small, srivastava2025towards}. Our understanding on how to design efficient and performant \textit{smaller} systems for domain-specific  remains limited, relying on opaque trial-and-error methods. 

%Pre-training SLMs from scratch: Architecture choice; Parameter Sharing; Data Filtering; Multiple-round training. Supervised Fine-tuning: Pretrain-then-finetune; Instruction tuning; Preference optimization. Knowledge Distillation: Data quality; Distribution mismatch; Domain gap

% falling within a range where the lower bound is the minimum size at which the model exhibits emergent abilities for a specialized task, and the upper bound is the largest size manageable within limited resource conditions

%option 2 - general framing 
%Yet despite this progress, our scientific understanding of how these models learn remains limited; Much of today’s model development still relies on opaque trial-and-error.

\footnotetext{\scriptsize\url{https://the-decoder.com/gpt-4-has-a-trillion-parameters}}

In this paper, we present \textbf{Pico}, a modular framework designed to help researchers develop SLMs in a more scientifically rigorous manner. \textbf{Pico} consists of two libraries: \texttt{pico-train}, which provides a lightweight, transparent training loop for language models; and \texttt{pico-analyze}, a complementary toolkit for analyzing their learning dynamics. Conceptually, \texttt{pico-train} provides the infrastructure for training and systematically checkpointing model states and activations, while \texttt{pico-analyze} offers the tools to compute learning dynamics metrics and comparisons on those checkpoints.

By bridging training and analysis in a single open-source ecosystem, \textbf{Pico} lowers the barrier for conducting reproducible, hypothesis-driven research on small language model development. To support controlled experimentation, we also release a set of baseline models trained under standardized conditions, the \textbf{\texttt{pico-decoder}} suite. The suite is a starting point for researchers to build on and compare against. Case studies in \cref{section:case-studies} demonstrate how \textbf{Pico} enables researchers to build language models in a hypothesis-driven way.

\section{\textbf{Pico}}

Unlike existing pretraining or interpretability stacks, \textbf{Pico} integrates modular training with built-in support for learning dynamics. \cref{tab:pico_comparison} highlights this key advantage.

On the training side, \texttt{pico-train} automatically logs detailed activations, gradients, and weights at checkpoint intervals and enables researchers to efficiently train models for controlled experiments.

On the analysis side, \texttt{pico-analyze} operates directly on these in-situ logs, applying flexible metrics and component abstractions to track learning dynamics as they unfold. %Its design encourages iterative experimentation, and its unopinionated architecture lets users plug in custom hypotheses and interventions.

In this section we provide a concise overview of the two \textbf{Pico} libraries: \texttt{pico-train} and \texttt{pico-analyze}. 

\subsection{\texttt{pico-train}: A Minimalist Approach to Model Training}

\texttt{pico-train} is a lightweight, transparent framework for training small- to medium-scale language models. Unlike many existing training libraries that prioritize efficiency at the cost of clarity, \texttt{pico-train} is designed to be simple, modular, and easy to modify, making it a flexible foundation for experimentation in language model research.

Out of the box, \texttt{pico-train} implements \texttt{pico-decoder}, a LLaMA-style transformer \citep{touvron2023llama} that incorporates key features of modern autoregressive language models, including Grouped Query Attention (GQA) \citep{ainslie2023gqa}, Rotary Position Embeddings (RoPE) \citep{su2024rope}, FlashAttention \citep{dao2022flashattention}, SwiGLU activations \citep{shazeer2020glu}, and RMSNorm \citep{zhang2019rms}. All components, except FlashAttention, are re-implemented from scratch in plain PyTorch \citep{paszke-etal-2019-pytorch}, with an emphasis on readability and documentation. %Future iterations of \textbf{Pico} will introduce additional architectures, such as \texttt{pico-diffusion} and \texttt{pico-statespace} models, all adhering to the same guiding principle: every \texttt{pico-*} model must be simple, well-documented, and serve as a clear base implementation for the given model architecture.

To ensure efficient multi-GPU and distributed training, \texttt{pico-train} is built on Lightning Fabric \citep{lightning2015fabric} -- a framework that, like \textbf{Pico}, prioritizes simplicity and flexibility. Lightning Fabric enables users to scale up training across multiple GPUs or nodes without introducing excessive abstractions and ensures that the core training logic remains easy to understand and modify.

A distinguishing feature of \texttt{pico-train} is its systematic checkpointing and version control system. It automatically saves:
\begin{itemize}
    \item \textbf{Model states in both PyTorch- and Hugging Face-compatible formats} \citep{wolf2019huggingface}. This dual-format checkpointing enables straightforward loading with vanilla PyTorch or integration into the Hugging Face ecosystem, facilitating downstream tasks such as fine-tuning, inference, or model sharing. Researchers can thus easily plug \texttt{pico-train} outputs into existing pipelines.% or community projects.

    \item \textbf{Intermediate activations and gradients.} At user-defined intervals, the library gathers layerwise activations and gradients from the forward and backward passes on the current training batch. Optionally, it can also capture these metrics from a fixed evaluation batch for consistent comparisons over training. Collecting these tensors at each checkpoint provides a granular record of how representations and gradient flows evolve over time.

    \item \textbf{Training data batch.} We save out the batch of training data that was used to extract the set of activations and gradients at a given point in training.

    \item \textbf{Evaluation results.} Users can define and record evaluation metrics (e.g., validation perplexity, accuracy) alongside model checkpoints.
\end{itemize}
\vspace{-0.2em}
All checkpoints are automatically uploaded and version-controlled on Hugging Face, ensuring that researchers can revisit any point in training to analyze how the model evolved over time. These structured checkpoints integrate seamlessly with \texttt{pico-analyze}, enabling learning dynamics research with minimal setup.

To simplify experimentation, we release a pre-tokenized, pre-chunked, and pre-shuffled version of Dolma \citep{soldaini2024dolma}, a large, open-source English dataset, on Hugging Face: \href{https://huggingface.co/datasets/pico-lm/pretokenized-dolma}{\textbf{\texttt{pretokenized-dolma}}}. This dataset removes preprocessing overhead, ensures consistency across runs, and supports streaming to reduce storage needs. Using it is optional; users can substitute their own data if they prefer. Details on our preprocessing steps are in \cref{app:pretokenized-dolma}.

By focusing on minimalism, modularity, and transparency, \texttt{pico-train} makes it easy to modify all aspects of the training pipeline. 

\subsection{\texttt{pico-analyze}: A General-Purpose Framework for Studying Learning Dynamics}

\texttt{pico-analyze} is a companion tool to \texttt{pico-train} designed to make analyzing learning dynamics seamless and reproducible. It directly integrates with the checkpoints saved by \texttt{pico-train} that include activations, and gradients and enables researchers to compute the learning dynamics of trained models.

At its core, \texttt{pico-analyze} follows a simple abstraction: it applies metrics to components. Metrics provide quantitative insights into various aspects of model behavior, while components define the specific model elements being analyzed. This design allows for flexible and fine-grained analysis of training dynamics.

\paragraph{Metrics.} Out of the box, \texttt{pico-analyze} supports a range of built-in metrics, including:
\begin{itemize}
    \item \textbf{Sparsity Measures}: \textit{Gini coefficient} \citep{hurley2009gini} and \textit{Hoyer metric} \citep{hoyer2004sparsity} gauge how concentrated the values of a matrix are near zero.

    \item \textbf{Rank-Based Metrics}: \textit{Proportional Effective Rank} \citep{diehl-martinez-etal-2024-tending} captures a matrix’s “effective dimensionality,” while \textit{Condition Number} evaluates its numerical stability.

    \item \textbf{Representation Similarity}: \textit{CKA} \citep{kornblith2019cka} and \textit{PWCCA} \citep{morcos2018pwcca} compare activation patterns across layers or checkpoints, revealing how internal representations evolve.
    
    \item \textbf{Norms}: \textit{Frobenius}, \textit{Nuclear}, and \textit{Infinity} norms measure the scale of a tensor, spotlighting issues such as vanishing or exploding parameters.
\end{itemize}

\paragraph{Components.} Metrics can be computed on different types of components: %, which fall into two categories: 
\begin{itemize} 
\item \textbf{Simple components}: Individual weight matrices, gradients, or activations from a single layer. 
\item \textbf{Compound components}: Higher-level structures that combine multiple model elements. One example is the OV circuit, which tracks how information flows in transformer models by combining the value and output projection matrices in self-attention layers \cite{elhage2021mathematical}. 
\end{itemize}

This two-step abstraction is designed for extensibility; new metrics and component types can be easily defined, allowing researchers to tailor analyses to specific hypotheses about language model learning. We view \texttt{pico-analyze} not as a static toolset, but as a foundation for community-driven interpretability research.  

\section{Case Studies}
\label{section:case-studies}

We illustrate how \textbf{Pico} enables systematic, hypothesis-driven experimentation through two case studies.%: (1) meta-learning pre-training, (2) low-rank adapter pre-training.

%that would otherwise require significant custom engineering

\subsection{MAML}

\begin{figure}[hb!]
\includegraphics[width=\linewidth]{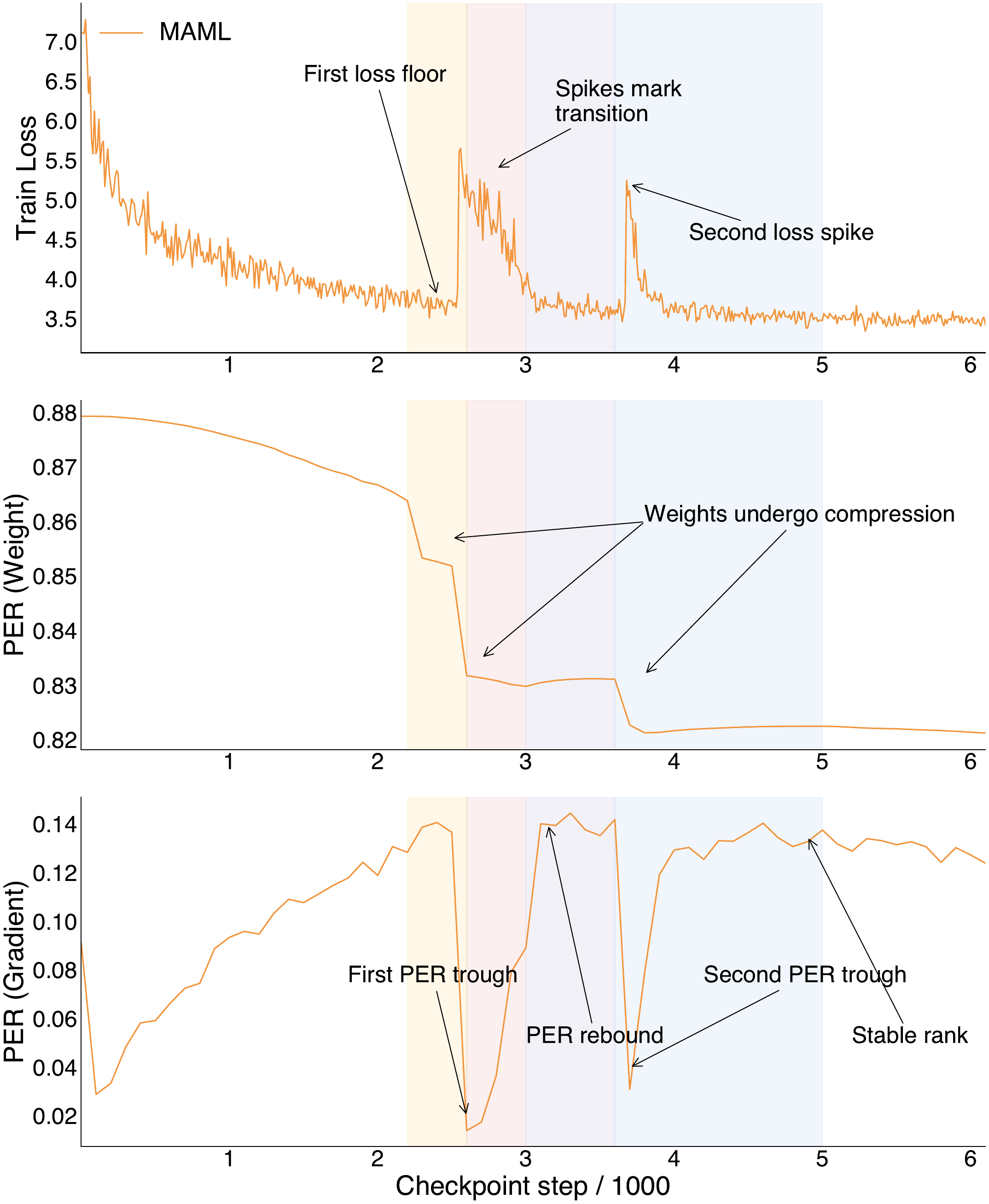}
\caption{
Training dynamics under MAML.
\textbf{Top to bottom:} Training loss and proportional effective rank (PER) of weights and gradients.
Sharp drops in PER align with spikes in loss. Shaded regions correspond to different observed phases in training. 
}
\label{fig:maml_example}
\end{figure}

Model-Agnostic Meta-Learning (\citealp[MAML]{pmlr-v70-finn17a}) trains models to adapt quickly by alternating between short bursts of task-specific learning and a global update that improves generalization. This setup encourages models to find initialization points that adapt well to new tasks. While MAML is typically used for fine-tuning, we follow prior work \citep{bansal-etal-2020-self, li-zhang-2021-semi} in applying it during pretraining.%, using synthetic token classification tasks.

\begin{table*}[htbp!]
\centering
\renewcommand{\arraystretch}{1.2}
\begin{tabular}{lcccccc}
\hline
\textbf{Family} & \textbf{Size} & \textbf{\#Tokens} &
\textbf{Paloma$\;\downarrow$} &
\textbf{HellaSwag$\;\uparrow$} &
\textbf{ARC-Easy$\;\uparrow$} &
\textbf{TruthfulQA$\;\uparrow$} \\
\hline\hline

\textbf{Pico} & 11M  & 250B & 136.17 & 25.62 & 32.79 & 51.75 \\
               & 65M  & 250B &  42.24 & 27.25 & 38.22 & 46.13 \\
               & 181M & 250B &  30.08 & 30.69 & 44.65 & 41.85 \\
               & 570M & 250B &  22.96 & 37.33 & 48.99 & 36.33 \\
\hline

Pythia & 14M  & 300B &  86.64 & 26.15 & 31.31 & 50.14 \\
                & 70M  & 300B &  43.76 & 27.56 & 36.23 & 47.02 \\
                & 160M & 300B &  29.96 & 30.26 & 43.73 & 44.51 \\
                & 410M & 300B &  20.55 & 40.55 & 52.10 & 41.23 \\
\hline

OPT    & 125M & 300B &  27.22 & 31.33 & 43.48 & 42.89 \\
                & 350M & 300B &  20.91 & 36.66 & 44.06 & 41.01 \\
\hline
\end{tabular}
\caption{Performance of small-scale language models on four benchmarks.
Lower is better for Paloma perplexity ($\downarrow$); higher is better for
HellaSwag, ARC-Easy, and TruthfulQA accuracies ($\uparrow$).}
\label{tab:model_benchmarks}
%\vspace{-0.5cm}
\end{table*}

\paragraph{Implementation} We implemented MAML in \texttt{pico-train} by adding a lightweight inner loop that updates a classification head on masked token tasks, followed by a meta-update to the full model \citep{africa2025meta,africa2025learning}. \texttt{pico-train} automatically handles distributed GPU synchronization, requiring no changes to \textbf{Pico}’s core training logic.

\paragraph{Analysis} We evaluate MAML on Paloma perplexity and observe consistent 4–15\% gains over standard pretraining. To better understand this improvement, we analyze the proportional effective rank (PER) \citep{diehl-martinez-etal-2024-tending} of both weights and gradients over time. PER captures the dimensionality of a tensor’s signal. In meta-learning, one concern is that inner-loop updates may overly constrain model updates to a low-dimensional subspace, potentially limiting generalization. As shown in \cref{fig:maml_example}, we observe synchronized troughs in PER and spikes in both loss and perplexity. This suggests that inner-loop updates temporarily compress the model’s capacity into a low-rank subspace before the outer-loop update restores variance and expressivity.

\paragraph{New Hypothesis and Next Steps}
These results support the hypothesis that MAML’s learning dynamics involve cycles of compression and recovery. This raises concrete follow-up questions: could adjusting the inner-loop learning rate reduce excessive compression? Would alternative meta-learning schedules or task mixes stabilize the representational space more effectively? Using Pico’s modular training and built-in logging, these variants can be tested with minimal friction. By comparing learning dynamics and outcomes across runs, researchers can refine their design choices in a reproducible, hypothesis-driven loop.% — illustrating Pico’s role as a practical sandbox for systematic small LM research.

%\paragraph{New Hypothesis} The analysis supports the hypothesis that MAML’s dynamics involve cycles of compression and recovery and raises several testable hypothesis, including: could tuning the inner-loop learning rate reduce the severity of PER compression? Would alternative meta-learning schedules stabilize representational variance?

%. Because \texttt{pico-analyze} stores and processes full training-time gradients, weights, and activations, it enables a microscope-level view into these dynamic shifts. %—surfacing subtle representational phenomena that would be missed with standard logging alone.

\subsection{ReLoRA}

\begin{figure}[ht!]
    \centering
    \includegraphics[width=\linewidth]{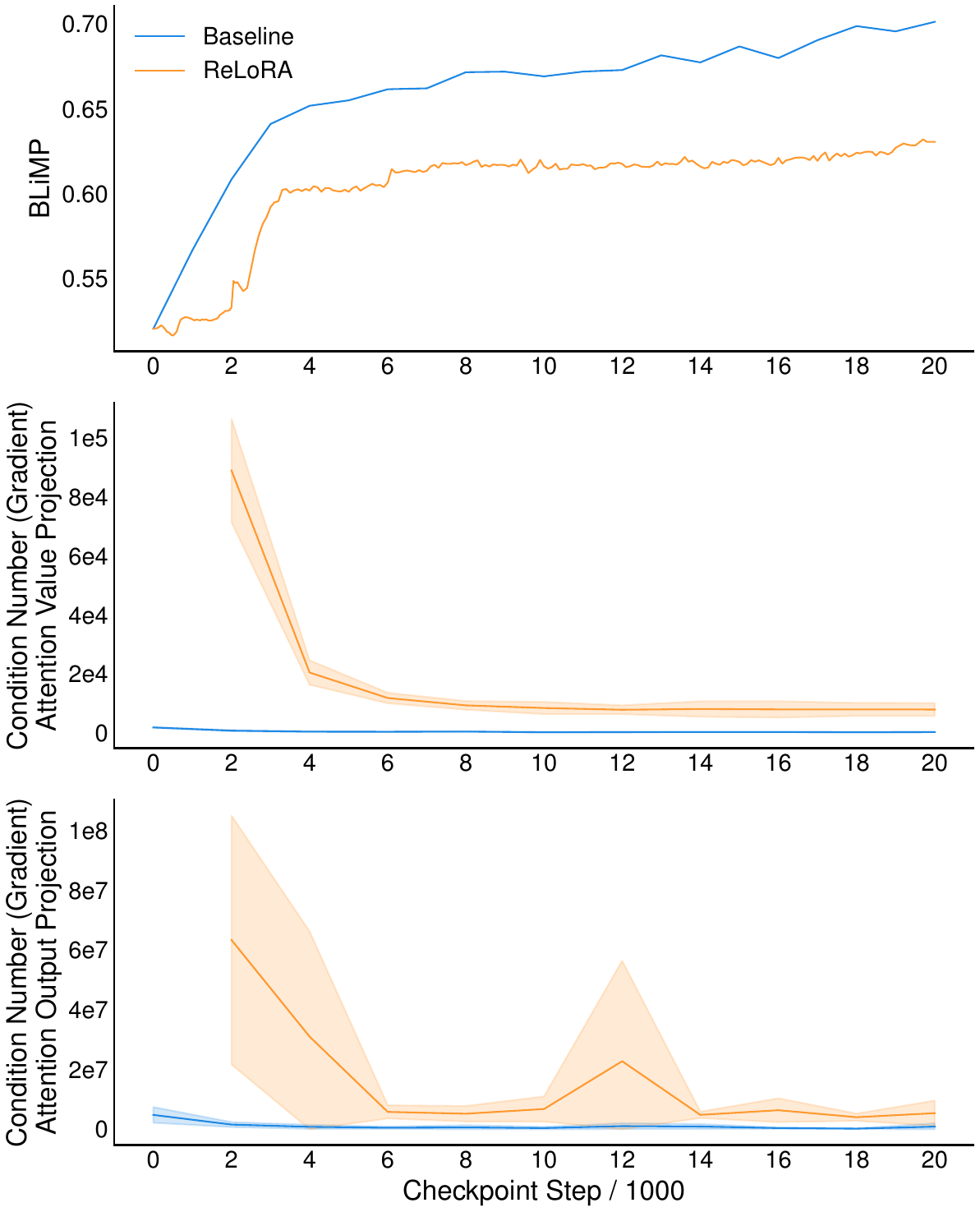}
    %\caption{The top portion shows training trajectories of BLiMP score, the bottom two plots the condition numbers of the gradient updates for the output and value projections in the attention mechanism. }
    \caption{
Training dynamics under ReLoRA.
\textbf{Top to bottom:} BLiMP accuracy over time, averaged condition numbers of the gradient updates for the attention value and output projection matrices. %Condition number shows wide inter-layer variance.
}
\label{fig:relora-graph}
\end{figure}

ReLoRA \citep{lialin_relora_2023} adapts LoRA \citep{hu_lora_2022}, a fine-tuning technique that freezes pretrained weights and injects trainable low-rank matrices, into the pretraining loop. In theory, this could provide a sample-efficient way to train large models by constraining updates to a low-rank subspace. %However, its impact on pretraining stability remains poorly understood.

\paragraph{Implementation} We incorporate ReLoRA into \texttt{pico-train} by adding a lightweight wrapper around attention and MLP weight matrices, and by modifying the learning rate schedule to handle periodic resets \citep{weiss2025investigatingreloraeffectslearning}. We evaluated the model on BLiMP \citep{warstadt_blimp_2020}, which we configured via a single config entry. % and required minimal code modification. 

\paragraph{Analysis} We find that ReLoRA surprisingly underperforms standard pretraining on BLiMP (see \cref{fig:relora-graph}, top) \citep{warstadt_blimp_2020}. To investigate this, we analyze the condition numbers of the gradient updates across layers and training checkpoints. This metric reflects how sensitive gradient-based updates are to numerical instability, a relevant concern for methods like ReLoRA that repeatedly project updates into a low-rank subspace. As shown in \cref{fig:relora-graph} (bottom), ReLoRA gradients are substantially more ill-conditioned and exhibit high inter-layer variance. 

% If the optimization geometry becomes poorly conditioned, learning can become more erratic.

\paragraph{New Hypothesis and Next Steps}
This pattern suggests that repeated low-rank resets may amplify gradient instability, undermining ReLoRA’s intended efficiency gains. Several next steps follow naturally: for example, adding layerwise condition number regularization, adjusting the rank dynamically, or modifying the reset schedule to reduce instability. Because \texttt{pico-train} and \texttt{pico-analyze} modularize core components and log detailed in-situ signals, researchers can test these changes quickly, track their effects on gradient stability, and iterate systematically. This demonstrates \textbf{Pico}’s value as a scientific sandbox for implementing, analyzing, and refining design choices in the small LM regime.

%\paragraph{New Hypothesis} This pattern suggests that repeated low-rank resets can amplify gradient instability, undermining the intended efficiency gains of ReLoRA during pretraining. With this insight, several hypotheses emerge for improving the method: for example, one could add layerwise condition number regularization.

%Because \texttt{pico-train} modularizes all core components this intervention could be implemented with minimal changes to the training loop. Combined with \texttt{pico-analyze}’s built-in metrics and checkpointing, researchers can rapidly iterate on these ideas, compare runs, and track how each modification affects gradient stability and downstream task performance. This exemplifies Pico’s role as a scientific sandbox for systematically testing improvements.

\section{\textbf{Pico} Model Suite}

We train a suite of \href{https://huggingface.co/pico-lm/models}{\textbf{\texttt{pico-decoder}}} models at various scales on \textbf{\texttt{pretokenized-dolma}} using \texttt{pico-train}, all of which are open-sourced on our Hugging Face organization. These models range from 11M to 570M parameters, with plans to extend to billion-parameter models, and serve both as evaluations of our training pipeline and as testbeds for research on scaling laws and interpretability. 

Each model is trained for 125,000 steps (covering 250B tokens). We evaluate the final model checkpoints on the Paloma benchmark \citep{magnusson2024paloma}, HellaSwag \citep{zellers2019hellaswag}, Arc-Easy \citep{clark2018arc} and Truthful QA \citep{lin2022truthfulqa}, comparing performance against established decoder models. As shown in \cref{tab:model_benchmarks}, our models achieve comparable results to Pythia and OPT models, despite running on an academic-level compute budget (4 nodes of 4 A100s each).

We provide a comparison of these models and their compute/storage overhead in \cref{tab:pico-decoder-configs}. Reported “GPU hours” are not directly comparable across frameworks due to differences in hardware, dataloaders, and logging pipelines, but our training times are broadly consistent with existing suites. For reference, whereas \texttt{pico-large} required 7,465 A100-hours, prior reports list Pythia-1B at 4,830 A100-hours, MPT-1.3B at 7,920 A100-hours, and TinyLlama-1.1B at 3,456 A100-hours under optimized stacks \citep{zhang2024tinyllama}. Importantly, all of our models are trained in a streaming pipeline, with datasets and checkpoints streamed from and uploaded to Hugging Face. Streaming adds data latency \footnote{On our network, streaming results in approximately 80–100\% slower batch loading compared to a local cache.} but greatly simplifies reproducibility and removes local storage requirements. Users who prioritize throughput can instead train models with a locally cached dataset.% by specifying this in the \texttt{pico-train} configuration.

\section{Related Literature}
\label{sec:related}

We survey where \textbf{Pico} sits within a growing ecosystem of frameworks that support the training and analysis of language models, ranging from optimized production libraries to interpretability toolkits. %and lightweight experimental suites.

\subsection{Training Frameworks}

%  and Databricks’ Dolly \cite{databricksdolly2023},
\paragraph{Open-source initiatives.} Initiatives by EleutherAI, including GPT-Neo, GPT-J, and the interpretability-focused Pythia suite \cite{biderman2023pythia} as well as projects like the Allen Institute's OLMo \citep{groeneveld2024olmo}, Meta’s Llama \cite{touvron2023llama}, and BigScience’s BLOOM \cite{le2023bloom}, have democratized access to pretrained weights and checkpoints. While these frameworks support post-hoc investigations into phenomena such as linguistic emergence and scaling effects \cite{belrose2023eliciting, gurnee2023finding, michaelov-bergen-2023-emergent, diehl-martinez-etal-2024-tending}, they do not capture detailed, in-training signals by default and only provide static checkpoints. Smaller frameworks like SmolLM \citep{allal2025smollm2}, TinyLlama \citep{zhang2024tinyllama}, NanoGPT \cite{karpathynanogpt2023},  and TinyStories \cite{eldan2023tinystoriessmalllanguagemodels} offer minimalistic, modular training loops that facilitate quick experimentation. However, they usually leave the implementation of fine-grained monitoring (e.g., activations or gradient flows) to the user.

\paragraph{Large-scale and efficient frameworks.} Platforms such as NVIDIA’s Megatron-LM \cite{narayanan2021efficient} and Microsoft’s DeepSpeed \cite{rasley2020deepspeed} excel at distributed training for models with billions of parameters, though they lack native mechanisms for inspecting intermediate states. 

\subsection{Analysis Frameworks}

\paragraph{Post-hoc model probing.} Given that detailed training signals are often unavailable by default, many researchers have adopted post-hoc probing methods. Such approaches rely on external hooking libraries to intercept hidden states and attention patterns \cite{voita2019analyzing, clark2019does, michel2019sixteen}, looking at information flows within models to discover security or privacy vulnerabilities \cite{roger2023largelanguagemodelsgenerate, YAO2024100211}. While powerful, these methods demand significant modifications and usually depend upon pre-existing checkpoints.% or training trajectory surrogates. 

\paragraph{Mechanistic interpretability.} Recently, mechanistic interpretability (mechinterp) has gained traction as a framework for reverse-engineering neural networks at the algorithmic level \cite{olah2020zoom, elhage2021mathematical}. Mechinterp focuses on localizing and characterizing the internal “circuitry” of attention heads, MLP layers, or individual neurons. A variety of mechinterp libraries, e.g., TransformerLens \cite{nanda2022transformerlens}, SAELens \cite{bloom2024saelens}, and ACDC \cite{conmy2023towards}, offer powerful tooling to dissect trained transformer models at inference time. However, these efforts typically assume that checkpoints are already available and do not natively capture the evolution of internal mechanisms throughout training.

\vspace{-0.5em}
\section{Conclusion}

We introduce \textbf{Pico}, a modular framework designed to help researchers study and improve small and medium-sized language models through a more systematic, scientifically grounded process. \texttt{pico-train} provides an extensible environment for training models, with built-in checkpointing that captures detailed signals needed to analyze learning dynamics. \texttt{pico-analyze} builds directly on these checkpoints, enabling researchers to test specific hypotheses about how changes to model architectures or training procedures affect convergence, sparsity, rank, and representation learning.

To further support controlled experiments and comparative studies, we open-source a suite of \textbf{\texttt{pico-decoder}} baseline models ranging from 11M to 570M parameters. These baselines give researchers a consistent starting point for evaluating new ideas or scaling laws under reproducible conditions.

By combining transparent training, detailed in-situ logging, and flexible analysis, \textbf{Pico} provides a practical sandbox for hypothesis-driven research. 

%We believe \textbf{Pico} will lower the barrier to systematic discovery and help move language model development from ad hoc trial-and-error toward a more rigorous science.

% We introduce \pico, a modular framework for training and analyzing language models. \texttt{pico-train} provides a minimal yet flexible environment for training language models that emphasizes transparency and checkpointing for learning dynamics research. \texttt{pico-analyze} uses these checkpoints to facilitate a broad set of analyses including  model convergence patterns, sparsity, and rank dynamics.

% Addtionally, we open-source a collection of \textbf{\texttt{pico-decoder}} models ranging from 11M to 570M parameters. These models are trained under controlled conditions, supporting research on scaling laws and representation learning.

% By bridging training and analysis in a lightweight, extensible framework, \textbf{Pico} lowers the barrier for studying language model learning dynamics. 

%Future work will expand the model suite and integrate advanced interpretability tools, further enhancing its utility for rigorous and reproducible research.

\section*{Limitations}

The \textbf{Pico} framework is designed for interpretability and experimentation rather than optimized large-scale production training, meaning it may not efficiently scale to industrial-scale models with hundreds of billions of parameters.  Additionally, the inherent overhead of systematically checkpointing intermediate activations and gradients at frequent intervals can significantly increase storage and computational costs during training.

\section*{Ethics Statement}

\textbf{Pico} aims to facilitate transparent and reproducible research into language model interpretability and learning dynamics. To streamline experimentation, we release the \textbf{\texttt{pretokenized-dolma}} dataset, a preprocessed dataset in English, enabling quick and efficient model training. Additionally, the initial \textbf{\texttt{pico-decoder}} model suite is also trained exclusively on English-language data. We acknowledge that this emphasis on English datasets and models can inadvertently reinforce English as the dominant language in NLP and interpretability research, potentially marginalizing research on other languages. We strongly encourage and support the development and release of similarly structured, high-quality datasets and models in languages other than English. Finally, any checkpoint or artifact uploaded by \textbf{Pico} to platforms such as Hugging Face must be used responsibly, with users remaining mindful of data privacy concerns, potential biases in training data, and risks associated with misuse or harmful applications of model checkpoints.

% =======================
% Acknowledgements
% =======================

\section*{Acknowledgments}

This work was supported by a grant from the Accelerate Programme for Scientific Discovery, made possible by a donation from Schmidt Futures. Richard Diehl Martinez is supported by the Gates Cambridge Trust (grant OPP1144 from the Bill \& Melinda Gates Foundation). Suchir Salhan is supported by Cambridge University Press \& Assessment. David Demitri Africa is supported by the Cambridge Trust and the Jardine Foundation. A big thank you to Leshem Choshen for their helpful comments and suggestions.

% =======================
% Acknowledgements
% =======================

\bibliography{custom}

% =======================
% Appendix
% =======================

\appendix
\label{sec:appendix}

\newpage
\onecolumn
\section{Default \texttt{pico-train} configurations}
\begin{table*}[h!]
    \centering
    \renewcommand{\arraystretch}{1} % Adjust row spacing
    \setlength{\tabcolsep}{8pt} % Adjust column spacing
    \begin{tabular}{|>{\centering\arraybackslash}p{3cm}|p{5cm}|p{6.5cm}|}
        \hline
        \textbf{Category} & \textbf{Parameter} & \textbf{Default Value} \\
        \hline
        \multirow{10}{*}{\textbf{Model}}  
            & Model Type & \texttt{pico\_decoder} \\
            & Hidden Dimension ($d_{\text{model}}$) & 768 \\
            & Number of Layers ($n_{\text{layers}}$) & 12 \\
            & Vocabulary Size & 50,304 \\
            & Sequence Length & 2,048 \\
            & Attention Heads & 12 \\
            & Key/Value Heads & 4 \\
            & Activation Hidden Dim & 3,072 \\
            & Normalization Epsilon & $1 \times 10^{-6}$ \\
            & Positional Embedding Theta & 10,000.0 \\
        \hline
        \multirow{7}{*}{\textbf{Training}}  
            & Optimizer & AdamW \\
            & Learning Rate & $3 \times 10^{-4}$ \\
            & LR Scheduler & Linear w/ Warmup \\
            & Warmup Steps & 2,500 \\
            & Gradient Accumulation Steps & 128 \\
            & Max Training Steps & 200,000 \\
            & Precision & BF16 Mixed \\
            %& Accelerator & CUDA \\
            %& Nodes & 1 \\
            %& Devices per Node & 1 \\
        \hline
        \multirow{3}{*}{\textbf{Data}}  
            & Dataset Name & \texttt{pico-lm/pretokenized-dolma} \\
            & Batch Size & 1,024 \\
            & Tokenizer & \texttt{allenai/OLMo-7B-0724-hf} \\
        \hline
        \multirow{6}{*}{\textbf{Checkpointing}}  
            & Auto Resume & True \\
            & Save Every N Steps & 1,000 \\
            %& Save to Hugging Face & False \\
            & Learning Dynamics Layers & \texttt{"attention.v\_proj",} \newline \texttt{"attention.o\_proj",} \newline \texttt{"swiglu.w\_2"} \\
            & Learning Dynamics Eval Data & \texttt{pico-lm/pretokenized-paloma-tinsy} \\
        \hline
        \multirow{3}{*}{\textbf{Evaluation}}  
            & Metrics & \texttt{["paloma"]} \\
            & Paloma Dataset Name & \texttt{pico-lm/pretokenized-paloma-tinsy} \\
            & Eval Batch Size & 16 \\
        \hline
        \multirow{3}{*}{\textbf{Monitoring}}  
            & Logging Level & INFO \\
            & Log Every N Steps & 100 \\
            %& Save to Weights \& Biases & False \\
        \hline
    \end{tabular}
    \caption{Default configuration settings used in \texttt{pico-train}, organized by configuration category.}
    \label{tab:default_configs}
\end{table*}

\vspace{-1em}
\section{\texttt{pico-decoder} models comparison}
\vspace{-1em}
\begin{table}[h!]
\centering
\renewcommand{\arraystretch}{1}
\begin{tabular}{|p{0.43\textwidth}||p{0.10\textwidth}|p{0.10\textwidth}|p{0.10\textwidth}|p{0.10\textwidth}|}
%\hline
%\multicolumn{5}{|c|}{\textbf{Pico-Decoder Model Comparison}} \\
\hline
\textbf{Attribute} & \texttt{tiny} & \texttt{small} & \texttt{medium} & \texttt{large} \\
\hline
Parameter Count & 11M & 65M & 181M & 570M \\
Hidden Dimension ($d_{\text{model}}$) & 96 & 384 & 768 & 1536 \\
Feed-forward Dim & 384 & 1536 & 3072 & 6144 \\
\hline
Training Time (125K Steps) & 4926h & 5645h & 6112h & 7465h \\
Checkpoint Time & 7m42s & 3m25s & 1m58s & 1m29s  \\
Checkpoint Storage (Model State) & 151MB & 863MB & 2.4GB & 7.5GB \\
Checkpoint Storage (Learning Dynamics) & 37MB & 176MB & 550MB & 2GB \\
\hline
\end{tabular}
\vspace{0.5em}
\caption{Comparison of \texttt{pico-decoder} model variants trained with default \texttt{pico-train} configurations. Except for hidden and feed-forward dimension, all models share the training settings detailed in \cref{tab:default_configs}. Models are trained for 125,000 total training steps on 16 NVIDIA A100-SXM4-80GB GPUs. Time reported in GPU hours (h), minutes (m) and seconds (s); checkpoint time and storage reported per checkpoint step.}
\label{tab:pico-decoder-configs}
\end{table}

\newpage
\onecolumn
\section{Available metrics in \texttt{pico-analyze}}

\begin{table*}[h!]
    \centering
    \renewcommand{\arraystretch}{1.1} % Adjust row spacing
    \setlength{\tabcolsep}{4pt}
    \begin{tabular}{|p{3.7cm}|p{7.3cm}|p{2.1cm}|p{1.9cm}|}
        \hline
        \textbf{Metric} & \textbf{Description} & \textbf{Data Type} & \textbf{Category} \\
        \hline
        \hline
        \textbf{CKA} \newline
        \citep{kornblith2019cka} &  
        \vspace{-0.8em}
        \begin{itemize}[leftmargin=*, itemsep=-1ex, topsep=0pt]
            \item Measures similarity between activations at different checkpoints
            \item Uses kernel methods to track representation evolution
            \vspace{-0.8em}
        \end{itemize}  
        & Activations & \textbf{Similarity} \\
        \hline
        \textbf{PWCCA} \newline \cite{morcos2018pwcca} & 
        \vspace{-0.8em}
        \begin{itemize}[leftmargin=*, itemsep=-1ex, topsep=0pt]
            \item Measures activation similarity across training
            \item Emphasizes important components via projections
            \vspace{-0.8em}
        \end{itemize}  
        & Activations & \textbf{Similarity} \\
        \hline
        \hline
        \textbf{Condition Number} &  
        \vspace{-0.8em}
        \begin{itemize}[leftmargin=*, itemsep=-1ex, topsep=0pt]
            \item Computes ratio of largest to smallest singular value
            \item Indicates sensitivity to small input changes
            \vspace{-0.8em}
        \end{itemize}
        & Weights\newline Activations\newline Gradients & \textbf{Rank} \\
        \hline
        \textbf{PER} \newline \citep{diehl-martinez-etal-2024-tending} &  
        \vspace{-0.8em}
        \begin{itemize}[leftmargin=*, itemsep=-1ex, topsep=0pt]
            \item Measures entropy of normalized singular values
            \item Indicates effective parameter usage
        \end{itemize}  
        & Weights\newline Gradients & \textbf{Rank} \\
        \hline
        \hline
        \textbf{Gini Coefficient} \newline \citep{hurley2009gini} &  
        \vspace{-0.8em}
        \begin{itemize}[leftmargin=*, itemsep=-1ex, topsep=0pt]
            \item Measures sparsity via weight distribution inequality
        \end{itemize}  
        & Weights\newline Activations\newline Gradients & \textbf{Sparsity} \\
        \hline
        \textbf{Hoyer's Sparsity} \newline \citep{hoyer2004sparsity} &  
        \vspace{-0.8em}
        \begin{itemize}[leftmargin=*, itemsep=-1ex, topsep=0pt]
            \item Measures sparsity by computing ratio of L1/L2 norms
        \end{itemize}  
        & Weights\newline Activations\newline Gradients & \textbf{Sparsity} \\
        \hline
        \hline
        \textbf{Norm} &  
        \vspace{-0.8em}
        \begin{itemize}[leftmargin=*, itemsep=-1ex, topsep=0pt]
            \item Frobenius, Nuclear, Infinity matrix norms
        \end{itemize}  
        & Weights\newline Activations\newline Gradients & \textbf{Norm} \\
        \hline
    \end{tabular}
    \caption{Overview of built-in metrics in \texttt{pico-analyze}. \textbf{Data Types} indicates on what types of checkpoint data the metrics can be applied. The \textbf{Category} column classifies metrics based on their primary purpose.}
    \label{tab:pico_analyze_metrics}
\end{table*}

\vspace{-1em}
\begin{multicols}{2} % <-- start two-column text environment here
\section{Preprocessing of the \texttt{pretokenized-dolma} dataset}
\label{app:pretokenized-dolma}
To prepare the \texttt{pretokenized-dolma} dataset used in our experiments, we begin by downloading the Dolma corpus and selecting a random 30\% subset. The text is then tokenized using the \verb|allenai/OLMo-7B-0724-hf| tokenizer and split into fixed-length sequences of 2049 tokens (2048 + 1 for next-token prediction). We ensure consistency across shards by chunking token streams without overlap, dropping any remainder shorter than the full sequence length.

After tokenization and chunking, we shuffle the dataset and sample a fixed number of sequences per shard, generating 100 shards in total. The resulting dataset is saved as Parquet files and uploaded to our Hugging Face organization under \href{https://huggingface.co/datasets/pico-lm/pretokenized-dolma}{\textbf{\texttt{pico-lm/pretokenized-dolma}}}.

To facilitate scalable loading and training, we further fine-shard the dataset into 10,000 pieces using a secondary script. These final shards are compact (78MB each), randomly shuffled, pre-tokenized, and ready for streaming via the Hugging Face datasets API. This preprocessing ensures that all models see data in a consistent order, which is critical for learning dynamics analysis. We release all of the scripts we use for preprocessing data in our GitHub repository.
\end{multicols} % <-- end two-column environment

\end{document}